\documentclass[11pt,a4paper]{article}
\usepackage[hyperref]{acl2020}
\usepackage{times}
\usepackage{latexsym}

\usepackage{latexsym}
\usepackage{amsmath}
\usepackage{amssymb}
\usepackage{bm}
\usepackage{bbold}
\usepackage{multirow}
\usepackage{subfig}
\usepackage{arydshln}
\usepackage{amsfonts}
\usepackage{bbm}
\usepackage{graphics}
\usepackage{standalone}
\usepackage{booktabs}
\usepackage{xspace}
\usepackage{enumitem}
\usepackage{color}
\usepackage{framed}

\usepackage{microtype}

\aclfinalcopy 

\newcommand{\ie}{\emph{i.e.,}\xspace}

\DeclareMathOperator*{\argmax}{arg\,max}

\title{How Does Selective Mechanism Improve Self-Attention Networks?}
  
\author{
Xinwei Geng$^1$\thanks{~~~Work done when interning at Tencent AI Lab.} ~~~ Longyue Wang$^2$ ~~~ Xing Wang$^2$ ~~~ Bing Qin$^{1,3}$ ~~~ Ting Liu$^{1,3}$ ~~~ Zhaopeng Tu$^2$ \\
$^1$Harbin Institute of Technology \quad\quad $^2$Tencent AI Lab \quad\quad $^3$Peng Cheng Laboratory\\
$^1$\{\tt xwgeng, qinb, tliu\}@ir.hit.edu.cn \\
$^2$\{\tt vinnylywang, brightxwang, zptu\}@tencent.com
}

\date{}

\begin{document}
\maketitle
\begin{abstract}
Self-attention networks (SANs) with selective mechanism has produced substantial improvements in various NLP tasks by concentrating on a subset of input words. However, the underlying reasons for their strong performance have not been well explained. 
In this paper, we bridge the gap by assessing the strengths of selective SANs (SSANs), which are implemented with a flexible and universal Gumbel-Softmax. 
Experimental results on several representative NLP tasks, including natural language inference, semantic role labelling, and machine translation, show that SSANs consistently outperform the standard SANs.
Through well-designed probing experiments, we empirically validate that the improvement of SSANs can be attributed in part to mitigating two commonly-cited weaknesses of SANs: {\em word order encoding} and {\em structure modeling}.
Specifically, the selective mechanism improves SANs by paying more attention to content words that contribute to the meaning of the sentence. The code and data are released at {\url{https://github.com/xwgeng/SSAN}}.
\end{abstract}

\section{Introduction}
Self-attention networks (SANs)~\cite{lin2017structured} have achieved promising progress in various natural language processing (NLP) tasks, including machine translation~\cite{Vaswani:2017:NIPS}, natural language inference~\cite{Shen:2018:AAAI}, semantic role labeling~\cite{tan2018deep,strubell2018linguistically} and language representation~\cite{devlin2019bert}. 
The appealing strength of SANs derives from high parallelism as well as flexibility in modeling dependencies among all the input elements. 

Recently, there has been a growing interest in integrating {\em selective mechanism} into SANs, which has produced substantial improvements in a variety of NLP tasks.
For example, 
some researchers incorporated a hard constraint into SANs to select a subset of input words, on top of which self-attention is conducted~\cite{Shen:2018:IJCAI,Hou:2019:SelectiveAB,Yang:2019:NAACL}.
\newcite{Yang:2018:EMNLP} and \newcite{guo2019gaussian} proposed a soft mechanism by imposing a learned Gaussian bias over the original attention distribution to enhance its ability of capturing local contexts. 
\newcite{Shen:2018:IJCAI} incorporated reinforced sampling to dynamically choose a subset of input elements, which are fed to SANs.

Although the general idea of selective mechanism works well across NLP tasks, previous studies only validate their own implementations in a few tasks, either on only classification tasks~\cite{Shen:2018:IJCAI,guo2019gaussian} or sequence generation tasks~\cite{Yang:2018:EMNLP,Yang:2019:NAACL}.
This poses a potential threat to the conclusive effectiveness of selective mechanism.
In response to this problem, we adopt a flexible and universal implementation of selective mechanism using Gumbel-Softmax~\cite{jang2016categorical}, called selective self-attention networks (\ie SSANs).
Experimental results on several representative types of NLP tasks, including natural language inference (i.e., {\em classification}), semantic role labeling (i.e., {\em sequence labeling}), and machine translation (i.e., {\em sequence generation}), demonstrate that SSANs consistently outperform the standard SANs (\S\ref{sec:downstream}).

Despite demonstrating the effectiveness of SSANs, the underlying reasons for their strong performance have not been well explained, which poses great challenges for further refinement.
In this study, we bridge this gap by assessing the strengths of selective mechanism on capturing essentially linguistic properties via well-designed experiments.
The starting point for our approach is recent findings: the standard SANs suffer from two representation limitation on modeling {\em word order encoding}~\cite{shaw2018self,yang:2019:assessing} and {\em syntactic structure modeling}~\cite{tang2018self,hao:2019:multi}, which are essential for natural language understanding and generation. 
Experimental results on targeted linguistic evaluation lead to the following observations:
\begin{itemize}
    \item SSANs can identify the improper word orders in both local (\S\ref{sec:local-order}) and global (\S\ref{sec:global-order}) ranges by learning to attend to the expected words.
    \item SSANs produce more syntactic representations (\S\ref{sec:syntactic:probing}) with a better modeling of structure by selective attention (\S\ref{sec:attention}).
    \item The selective mechanism improves SANs by paying more attention to content words that posses semantic content and contribute to the meaning of the sentence (\S\ref{sec:linguistic}).  
\end{itemize}

\section{Methodology}

\subsection{Self-Attention Networks}
SANs~\cite{lin2017structured}, as a variant of attention model~\cite{bahdanau2015neural,luong2015effective}, compute attention weights between each pair of elements in a single sequence. 
Given the input layer ${\bf H} = \{{\bf h}_1,\cdots, {\bf h}_N\} \in \mathbbm{R}^{N \times d}$, SANs first transform the layer $\bf H$ into the queries ${\bf Q} \in \mathbbm{R}^{N \times d}$, the keys ${\bf K} \in \mathbbm{R}^{N \times d}$, and the values ${\bf V} \in \mathbbm{R}^{N \times d}$ with three separate weight matrices. The output layer $\bf O$ is calculated as:
\begin{equation}
    {\bf O} = \textsc{ATT}({\bf Q}, {\bf K}) {\bf V}
    \label{eqn:san}
\end{equation}
where the alternatives to $\textsc{ATT}(\cdot)$ can be additive attention~\cite{bahdanau2015neural} or dot-product attention~\cite{luong2015effective}. Due to time and space efficiency, we used the dot-product attention in this study, which is computed as:
\begin{equation}\label{eq:2}
    \textsc{ATT}({\bf Q}, {\bf K}) = softmax(\frac{{\bf Q}{\bf K}^{T}}{\sqrt{d}})
\end{equation}
where $\sqrt{d}$ is the scaling factor with $d$ being the dimensionality of layer states~\cite{Vaswani:2017:NIPS}.

\subsection{Weaknesses of Self-Attention Networks}

Despite SANs have demonstrated its effectiveness on various NLP tasks, recent studies empirically revealed that SANs suffer from two representation limitations of modeling {\em word order encoding} \cite{yang:2019:assessing} and {\em syntactic structure modeling} \cite{tang2018self}. In this work, we concentrate on these two commonly-cited issues.

\paragraph{Word Order Encoding}
SANs merely rely on attention mechanism with neither recurrence nor convolution structures. In order to incorporate sequence order information, \newcite{Vaswani:2017:NIPS} proposed to inject position information into the input word embedding with additional position embedding. Nevertheless, SANs are still weak at learning word order information~\cite{yang:2019:assessing}. 
Recent studies have shown that incorporating recurrence~\cite{Chen:2018:ACL,Hao-2019-towards,hao2019modeling}, convolution~\cite{song-etal-2018-double,Yang:2019:NAACL}, or advanced position encoding~\cite{shaw2018self,Wang:2019:EMNLP} into vanilla SANs can further boost their performance, confirming its shortcomings at modeling sequence order.

\paragraph{Structure Modeling}
Due to lack of supervision signals of learning structural information, recent studies pay widespread attention on incorporating syntactic structure into SANs. For instance, \newcite{strubell2018linguistically} utilized one attention head to learn to attend to syntactic parents of each word. 
Towards generating better sentence representations, several researchers propose phrase-level SANs by performing self-attention across words inside a n-gram phrase or syntactic constituent~\cite{wu-etal-2018-phrase,hao:2019:multi,wang:2019:tree}.
These studies show that the introduction of syntactic information can achieve further improvement over SANs, demonstrating its potential weakness on structure modeling.

\subsection{{\em Selective} Self-Attention Networks}

\begin{figure}[t]
\centering
\includegraphics[width=0.48\textwidth]{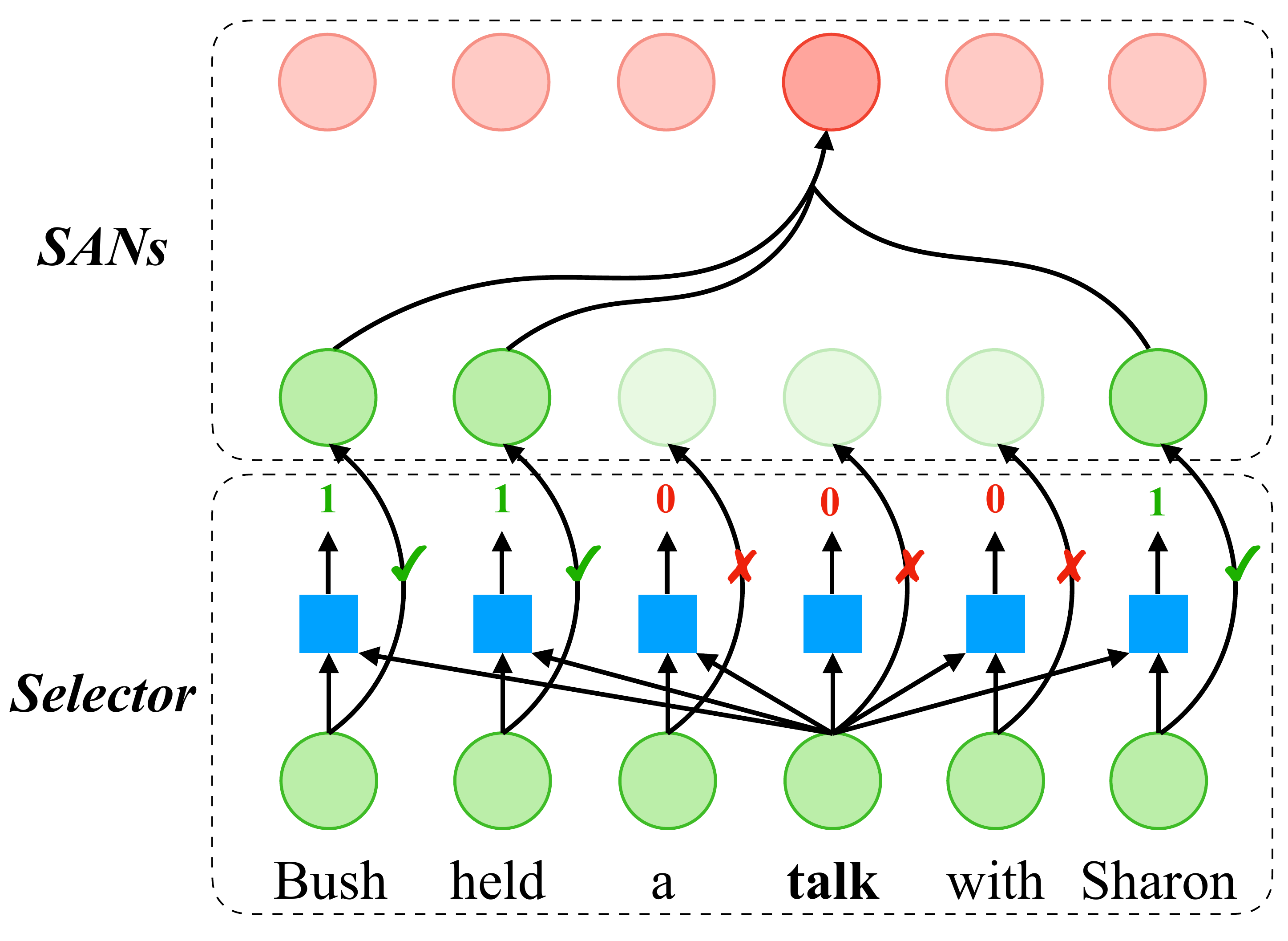}
\caption{Illustration of SSANs that select a subset of input elements with an additional selector network, on top of which self-attention is conducted.
In this example, the word ``{\em talk}'' performs attention operation over input sequence, where the words ``{\em Bush}'', ``{\em held}'' and ``{\em Sharon}'' are chosen as the truly-significant words.}
\label{fig:cls}
\end{figure}

In this study, we implement the selective mechanism on SANs by introducing an additional {\em selector}, namely SSANs, as illustrated in Figure~\ref{fig:cls}. The selector aims to select a subset of elements from the input sequence, on top of which the standard self-attention (Equation~\ref{eqn:san}) is conducted. We implement the selector with Gumbel-Softmax, which has proven effective for computer vision tasks \cite{shen:2018:sharp,yang:2019:modeling}.

\paragraph{Selector}
Formally, we parameterize selection action $a \in \{\textsc{SELECT}, \textsc{DISCARD}\}$ for each input element with an auxiliary policy network, where \textsc{SELECT} indicates that the element is selected for self-attention while \textsc{DISCARD} represents to abandon the element. The output action sequence ${\bf A} \in \mathbbm{R}^{N}$ is calculated as:
\begin{align}
    \pi({\bf A}) &= sigmoid({\bf E}_s) \\
    {\bf E}_s &= {\bf Q}_s{\bf K}_s^T
\end{align}
where ${\bf Q}_s  \in \mathbbm{R}^{N \times d}$ and ${\bf K}_s \in \mathbbm{R}^{N \times d}$ are transformed from the input layer ${\bf H}$ with distinct weight matrices. We utilize $sigmoid$ as activation function to calculate the distribution for choosing the action \textsc{SELECT} with the probability $\pi$ or \textsc{DISCARD} with the probability $1 - \pi$.

\paragraph{Gumbel Relaxation}
There are two challenges for training the selector: (1) the ground-truth labels indicating which words should be selected are unavailable; and (2) the discrete variables in ${\bf A}$ lead to a non-differentiable objective function. In response to this problem, \newcite{jang2016categorical} proposed Gumbel-Softmax to give a continuous approximation to sampling from the categorical distribution. We adopt a similar approach by adding Gumbel noise~\cite{gumbel1954statistical} in the sigmoid function, which we refer as {\em Gumbel-Sigmoid}. Since sigmoid can be viewed as a special 2-class case (${\bf E}_s$ and ${\bf 0}$ in our case) of softmax, we derive the {\em Gumbel-Sigmoid} as:
\begin{equation}
    \begin{split}
    \text{\em Gum}&\text{\em bel-Sigmoid}({\bf E}_s) \\
    &= \text{\em sigmoid}(({\bf E}_s + {\bf G}^{\prime} - {\bf G}^{\prime\prime}) / \tau)\\
    &= \frac{\exp(({\bf E}_s + {\bf G}^{\prime})/\tau)}{\exp(({\bf E}_s + {\bf G}^{\prime})/\tau) + \exp({\bf G}^{\prime\prime}/\tau)}\\
    \end{split}
\end{equation}
where ${\bf G}^{\prime}$ and ${\bf G}^{\prime\prime}$ are two independent Gumbel noises~\cite{gumbel1954statistical}, and $\tau \in (0, \infty)$ is a temperature parameter. As $\tau$ diminishes to zero, a sample from the {\em Gumbel-Sigmoid} distribution becomes cold and resembles the one-hot samples. 
At training time, we can use {\em Gumbel-Sigmoid} to obtain differentiable sample ${\bf A}$ as $\text{\em Gumbel-Sigmoid}({\bf E}_s)$. 
In inference, we choose the action with maximum probability as the final output.

\section{NLP Benchmarks}
\label{sec:downstream}

To demonstrate the robustness and effectiveness of the SSANs, we evaluate it in three representative NLP tasks: language inference, semantic role labeling and machine translation. We used them as NLP benchmarks, which cover classification, sequence labeling and sequence generation categories.
Specifically, the performances of semantic role labeling and language inference models heavily rely on structural information~\cite{strubell2018linguistically}, while machine translation models need to learn word order and syntactic structure~\cite{Chen:2018:ACL,hao2019modeling}.

\subsection{Experimental Setup}

\paragraph{Natural Language Inference} aims to classify semantic relationship between a pair of sentences, \ie a premise and corresponding hypothesis.
We conduct experiments on the Stanford Natural Language Inference (SNLI) dataset~\cite{snli:emnlp2015}, which has three classes: Entailment, Contradiction and Neutral. 

We followed \newcite{Shen:2018:AAAI} to use a token2token SAN layer followed by a source2token SAN layer to generate a compressed vector representation of input sentence. The selector is integrated into the token2token SAN layer. 
Taking the premise representation $s^p$ and the hypothesis vector $s^h$ as input, their semantic relationship is represented by the concatenation of $s^p$, $s^h$, $s^p - s^h$ and $s^p \cdot s^h$, which is passed to a classification module to generate a categorical distribution over the three classes. 
We initialize the word embeddings with 300D GloVe 6B pre-trained vectors~\cite{pennington2014glove}, and the hidden size is set as 300.

\paragraph{Semantic Role Labeling}\label{sec:srl}
is a shallow semantic parsing task, which aims to recognize the predicate-argument structure of a sentence, such as ``who did what to whom'', ``when'' and ``where''. Typically, it assigns labels to words that indicate their semantic role in the sentence.
Our experiments are conducted on CoNLL2012 dataset provided by \newcite{toward:2013:conll}. 

We evaluated selective mechanism on top of DEEPATT\footnote{\url{https://github.com/XMUNLP/Tagger}.}~\cite{tan2018deep}, which consists of stacked SAN layers and a following softmax layer. Following their configurations, we set the number of SAN layers as 10 with hidden size being 200, the number of attention heads as 8 and the dimension of word embeddings as 100.
We use the GloVe embeddings~\cite{pennington2014glove}, which are pre-trained on Wikipedia and Gigaword, to initialize our networks, but they are not fixed during training. We choose the better feed-forward networks (FFN) variants of DEEPATT as our standard settings.

\paragraph{Machine Translation}\label{sec:mt} is a conditional generation task, which aims to translate a sentence from a source language to its counterpart in a target language.
We carry out experiments on several widely-used datasets, including small English$\Rightarrow$Japanese (En$\Rightarrow$Ja) and English$\Rightarrow$Romanian (En$\Rightarrow$Ro) corpora, as well as a relatively large English$\Rightarrow$German (En$\Rightarrow$De) corpus. 
For En$\Rightarrow$De and En$\Rightarrow$Ro, we respectively follow \citet{Li:2018:EMNLP} and \citet{He:Layer:NIPS} to prepare WMT2014\footnote{\url{http://www.statmt.org/wmt14}.} and IWSLT2014\footnote{\url{https://wit3.fbk.eu/mt.php?release=2014-01}.} corpora. For En$\Rightarrow$Ja, we use KFTT\footnote{\url{http://www.phontron.com/kftt}.} dataset provided by \citet{neubig11kftt}.
All the data are tokenized and then segmented into subword symbols using BPE~\cite{sennrich2015neural} with 32K operations.

We implemented the approach on top of advanced \textsc{Transformer} model~\cite{Vaswani:2017:NIPS}.
On the large-scale En$\Rightarrow$De dataset, we followed the base configurations to train the NMT model, which consists of 6 stacked encoder and decoder layers with the layer size being 512 and the number of attention heads being 8. 
On the small-scale En$\Rightarrow$Ro and En$\Rightarrow$Ja datasets, we followed~\newcite{He:Layer:NIPS} to decrease the layer size to 256 and the number of attention heads to 4. 

For all the tasks, we applied the selector to the first layer of encoder to better capture lexical and syntactic information, which is empirically validated by our further analyses in Section~\ref{sec:order}.

\subsection{Experimental Results}

\begin{table}[t]
  \centering
  \begin{tabular}{ c || r || c c | r}
    {\bf Task} & {\bf Size} & {\bf SANs} & {\bf SSANs} & {\bf $\bigtriangleup$} \\
    \hline\hline
    \multicolumn{5}{c}{{Natural Language Inference} (Accuracy)} \\
    \hline
    SNLI & 550K & 85.60 & 86.30 & {\em +0.8\%}\\
    \hline
    \multicolumn{5}{c}{Semantic Role Labeling (F1 score)} \\
    \hline
    CoNLL & 312K & 82.48 & 82.88 & {\em +0.5\%}\\
    \hline
    \multicolumn{5}{c}{{Machine Translation} (BLEU)} \\
    \hline
  		 En$\Rightarrow$Ro & 0.18M & 23.22 & 23.91  & {\em +3.0\%} \\
         En$\Rightarrow$Ja & 0.44M & 31.56 & 32.17 & {\em +1.9\%} \\
         En$\Rightarrow$De & 4.56M & 27.60 & 28.50 & {\em +3.3\%}\\
  \end{tabular}
  \caption{Results on the NLP benchmarks. ``Size'' indicates the number of training examples, and ``$\bigtriangleup$'' denotes relative improvements over the vanilla SANs.} 
  \label{tab:benchmark}
\end{table}

Table~\ref{tab:benchmark} shows the results on the three NLP benchmarks. Clearly, introducing selective mechanism significantly and consistently improves performances in all tasks, demonstrating the universality and effectiveness of the selective mechanism for SANs. 
Concretely, SSANs relatively improve prediction accuracy over SANs by +0.8\% and +0.5\% respectively on the NLI and SRL tasks, showing their superiority on structure modeling. \citet{Shen:2018:IJCAI} pointed that SSANs can better capture dependencies among semantically important words, and our results and further analyses (\S\ref{sec:structure}) provide supports for this claim.

In the machine translation tasks, SSANs consistently outperform SANs across language pairs. Encouragingly, the improvement on translation performance can be maintained on the large-scale training data. The relative improvements on the En$\Rightarrow$Ro, En$\Rightarrow$Ja, and En$\Rightarrow$De tasks are respectively +3.0\%, +1.9\%, and +3.3\%.
We attribute the improvement to the strengths of SSANs on word order encoding and structure modeling, which are empirically validated in Sections~\ref{sec:order} and~\ref{sec:structure}.

\newcite{Shen:2018:IJCAI} implemented the selection mechanism with the REINFORCE algorithm. \citet{jang2016categorical} revealed that compared with Gumbel-Softmax~\cite{Maddison:NIPS:sampling}, REINFORCE~\cite{Williams1992Simple} suffers from high variance, which consequently leads to slow converge. 
In our preliminary experiments, we also implemented {REINFORCE}-based SSANs, but it underperforms the Gumbel-Softmax approach on the benchmarking En$\Rightarrow$De translation task (BLEU: 27.90 vs. 28.50, not shown in the paper). 
The conclusion is consistent with \newcite{jang2016categorical}, and we thus use Gumbel-Softmax instead of REINFORCE in this study.

\section{Evaluation of Word Order Encoding}
\label{sec:order}

In this section, we investigate the ability of SSANs of capturing both {\em local} and {\em global} word orders on the {\em bigram order shift detection} (\S\ref{sec:local-order}) and {\em word reordering detection} (\S\ref{sec:global-order}) tasks.

\subsection{Detection of Local Word Reordering}
\label{sec:local-order}

\paragraph{Task Description} 
~\newcite{conneau2018you} propose a {\em bigram order shift detection} task to test whether an encoder is sensitive to local word orders. 
Given a monolingual corpus, a certain portion of sentences are randomly extracted to construct instances with illegal word order. Specially, given a sentence $X = \{x_1, \dots, x_N\}$, two adjacent words (\ie $x_n$, $x_{n+1}$) are swapped to generate an illegal instance $X^{\prime}$ as a substitute for $X$. 
Given processed data which consists of intact and inverted sentences, examined models are required to distinguish intact sentences from inverted ones. To detect the shift of bigram word order, the models should learn to recognize normal and abnormal word orders.

The model consists of 6-layer SANs and 3-layer MLP classifier. The layer size is 128, and the filter size is 512.
We trained the model on the open-source dataset\footnote{\url{https://github.com/facebookresearch/SentEval/tree/master/data/probing}.} provided by \newcite{conneau2018you}.
The accuracy of SAN-based encoder is higher than previously reported result on the same task~\cite{Li:2019:NAACL} (52.23 vs. 49.30).

\begin{table}[t]
  \centering
  \begin{tabular}{ l | c || c  | r}
      {\bf Model} & \bf {Layer} & {\bf Acc.} & {$\bigtriangleup$} \\
     \hline
     {\bf SANs} & -- & 52.23 & --\\
     \hline
     \multirow{6}{*}{{\bf SSANs}} & 1 & {\bf 62.55} & {\em +19.8\%} \\
     & 2 & 53.73 & {\em +2.9\%} \\
     & 3 & 54.65 & {\em +4.6\%} \\
     & 4 & 54.29 & {\em +3.9\%} \\
     & 5 & 54.78 & {\em +4.9\%} \\
     & 6 & 54.23 & {\em +3.8\%} \\
  \end{tabular}
  \caption{Results on the {\em local} {bigram order shift detection} task when {\em SSANs} are applied into different layers.}
  \label{tab:order}
\end{table}

\paragraph{Detection Accuracy}
Table~\ref{tab:order} lists the results on the local bigram order shift detection task, in which SSANs are applied to different encoder layers. Clearly, all the SSANs variants consistently outperform SANs, demonstrating the superiority of SSANs on capturing local order information. Applying the selective mechanism to the first layer achieves the best performance, which improves the prediction accuracy by +19.8\% over SANs.
The performance gap between the SSANs variants is very large (i.e., 19.8\% vs. around 4\%), which we attribute to that the detection of local word reorder depends more on lexical information embedded in the bottom layer.

\begin{figure}[t]
    \centering
    \includegraphics[width=0.4\textwidth]{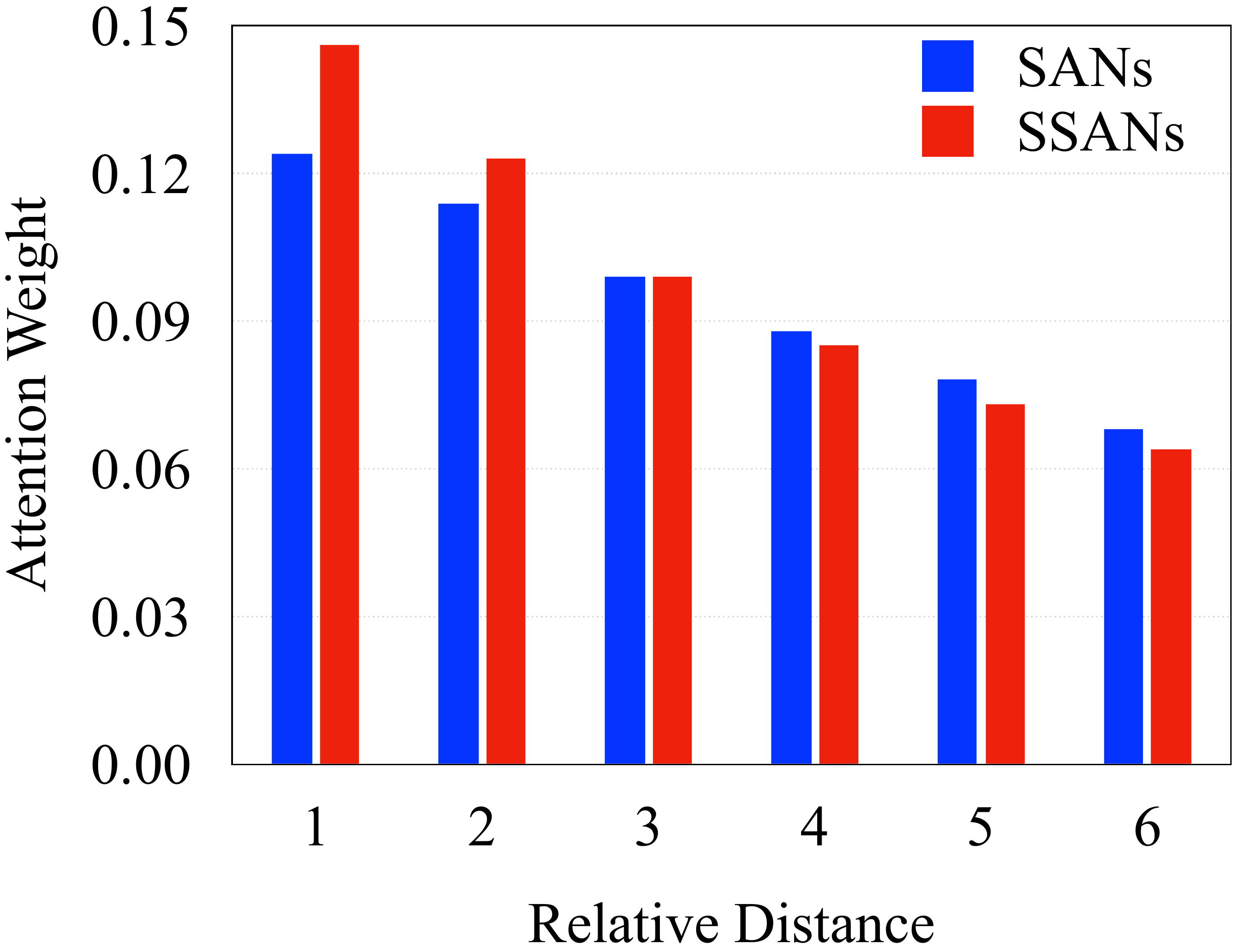}
	\caption{Attention weights over attended words with different relative distance from the query word on the {\em local} reordering task.
	{\em SSANs pay more attention to the adjacent words (distance=1) than SANs}.} 
	\label{fig:order:distance}
\end{figure}

\begin{figure}[t]
    \begin{center}
        \subfloat[SANs]{\includegraphics[height=0.23\textwidth]{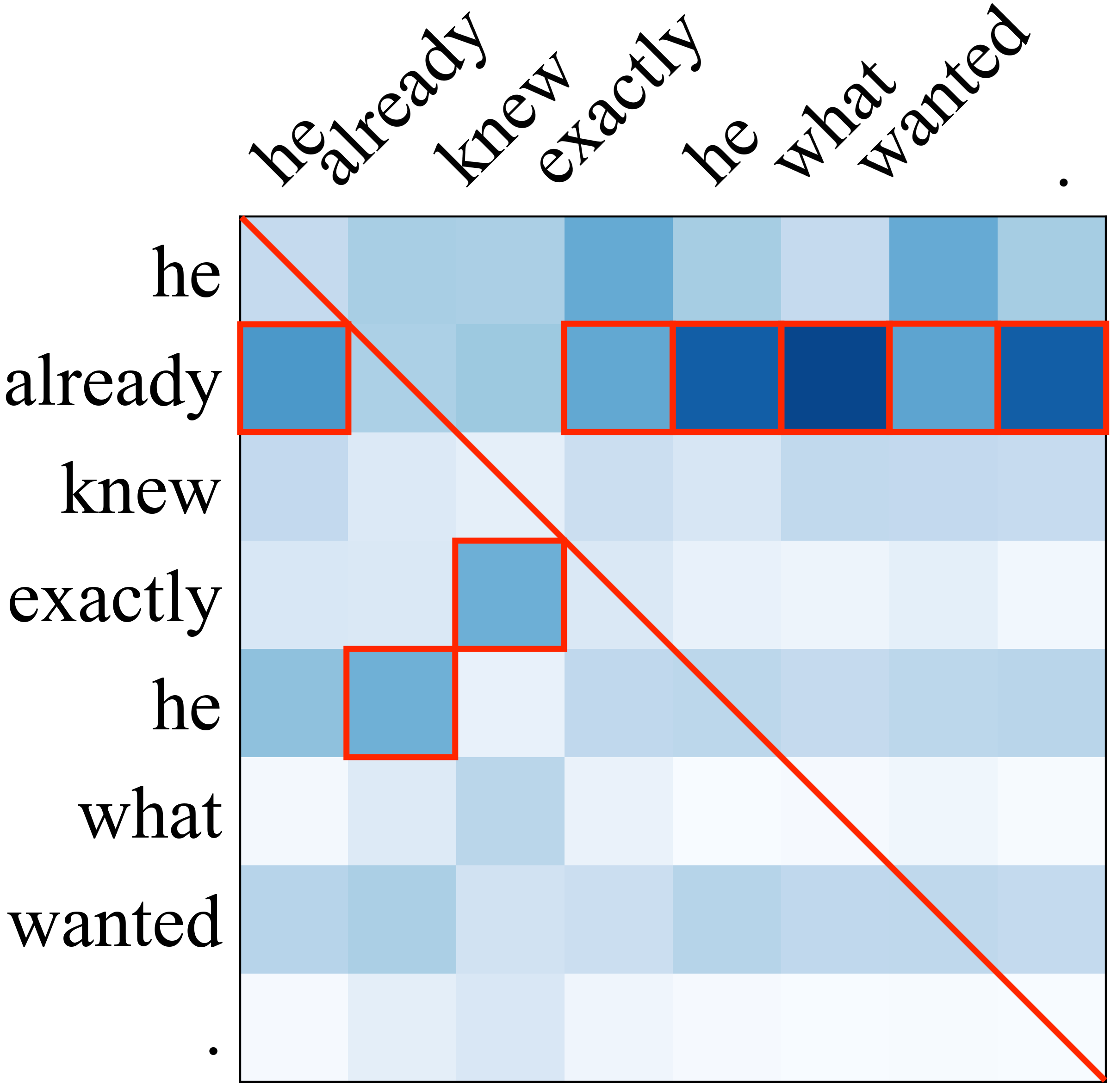}}
        \hfill
        \subfloat[SSANs]{\includegraphics[height=0.23\textwidth]{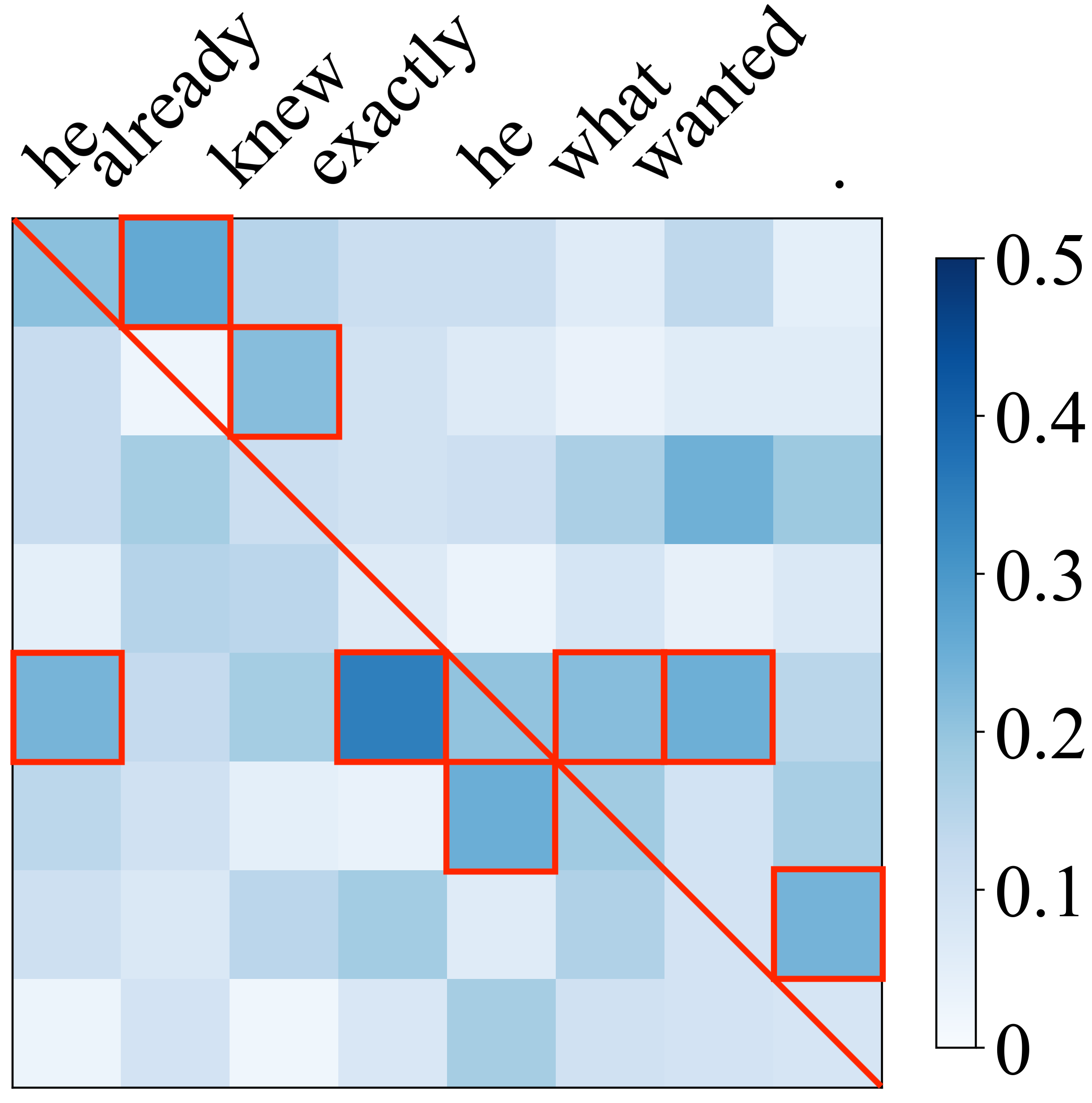}}
    \end{center}
	\caption{Visualization of attention weights from an example on the {\em local} reordering detection task. We highlight the attended word (Y-axis) with maximum attention weight for each query (X-axis) in red rectangles.} 
	\label{fig:order:case}
\end{figure}

\paragraph{Attention Behaviors}
The objective of local reordering task is to distinguish the swap of two adjacent words, which requires the examined model to pay more attention to the adjacent words.
Starting from this intuition, we investigate the attention distribution over the attended words with different relative distances from the query word, as illustrated in Figure~\ref{fig:order:distance}. We find that both SANs and SSANs focus on neighbouring words (e.g., distance $<$ 3), and SSANs pays more attention to the adjacent words (distance=1) than SANs (14.6\% vs. 12.4\%). The results confirm our hypothesis that the selective mechanism helps to exploit more bigram patterns to accomplish the task objective. 
Figure~\ref{fig:order:case} shows an example, in which SSANs attend most to the adjacent words except the inverted bigram ``{\em he what}''. 
In addition, the surrounding words ``{\em exactly}'' and ``{\em wanted}'' also pay more attention to the exceptional word ``{\em he}''. We believe such features help to distinguish the abnormally local word order.

\subsection{Detection of Global Word Reordering}
\label{sec:global-order}

\paragraph{Task Description} 
\newcite{yang:2019:assessing} propose a {\em word reordering detection} task to investigate the ability of SAN-based encoder to extract global word order information.
Given a sentence $X = \{x_1, \dots, x_N \}$,  a random word $x_i$ is popped and inserted into another position $j$ ($i \neq j$). The objective is to detect both the original position the word is popped out (labeled as ``O''), and the position the word is inserted (labeled as ``I'').

The model consists of 6-layer SANs and a output layer. The layer size is 512, and the filter size is 2048.
We trained the model on the open-source dataset\footnote{\url{https://github.com/baosongyang/WRD}.} provided by~\newcite{yang:2019:assessing}.

\begin{table}[t]
  \centering
  \begin{tabular}{ c | c || c c |c}
      {\bf Model} & \bf {Layer} & {\bf Insert} &{\bf Original} & {\bf Both} \\
     \hline
     {\bf SANs} & -- & 73.20 & 66.00 & 60.10\\
     \hline
     \multirow{6}{*}{{\bf SSANs}} & 1 & {\bf 81.52} & {\bf 72.19} & {\bf 66.77}\\
     & 2 & 80.14 & 70.01 & 63.97 \\
     & 3 & 79.82 & 69.69 & 63.93\\
     & 4 & 79.08 & 70.22 & 63.67\\
     & 5 & 80.19 & 69.84 & 64.12 \\
     & 6 & 80.27 & 69.50 & 63.73\\
  \end{tabular}
  \caption{Performance on the {\em global} {word reordering detection} (WRD) task.} 
  \label{tab:reorder}
\end{table}

\paragraph{Detection Accuracy}
Table~\ref{tab:reorder} lists the results on the global reordering detection task, in which all the SSANs variants improve prediction accuracy. Similarly, applying the selective mechanism to the first layer achieves the best performance, which is consistent with the results on the global word reordering task (Table~\ref{tab:order}).
However, the performance gap between the SSANs variants is much lower that that on the local reordering task (i.e., 4\% vs. 15\%). 
One possible reason is that the detection of global word reordering may also need syntactic and semantic information, which are generally embedded in the high-level layers~\cite{Peters:2018:NAACL}.

\begin{figure}[t]
    \centering
    \includegraphics[width=0.4\textwidth]{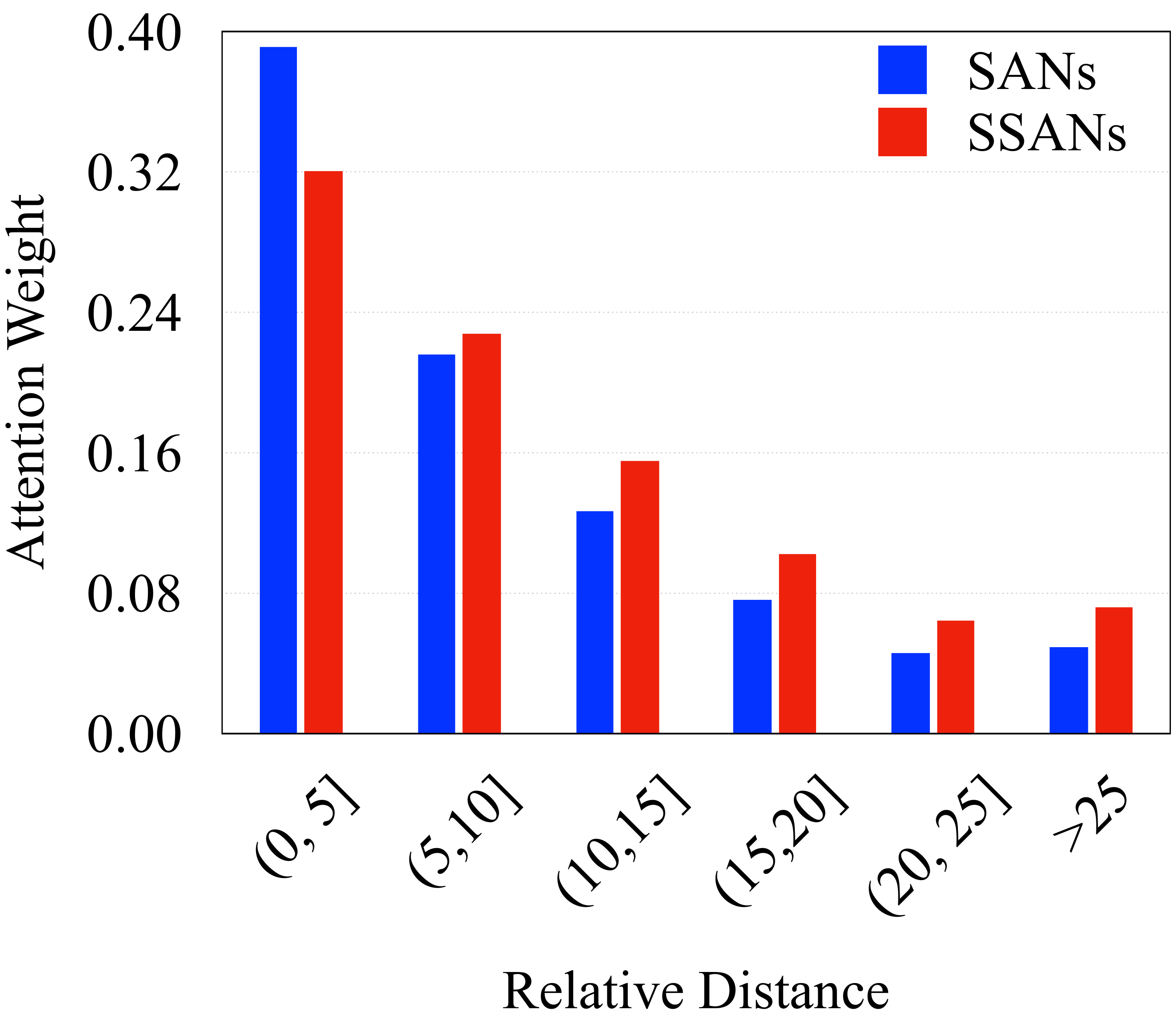}
	\caption{Attention weights over attended words with different relative distance from the query word on the {\em global} WRD task. {\em SSANs pay more attention to the distant words (distance$>5$) than SANs}.}
	\label{fig:reorder:distance}
\end{figure}

\begin{figure}[t]
    \begin{center}
        \subfloat[SANs]{\includegraphics[height=0.23\textwidth]{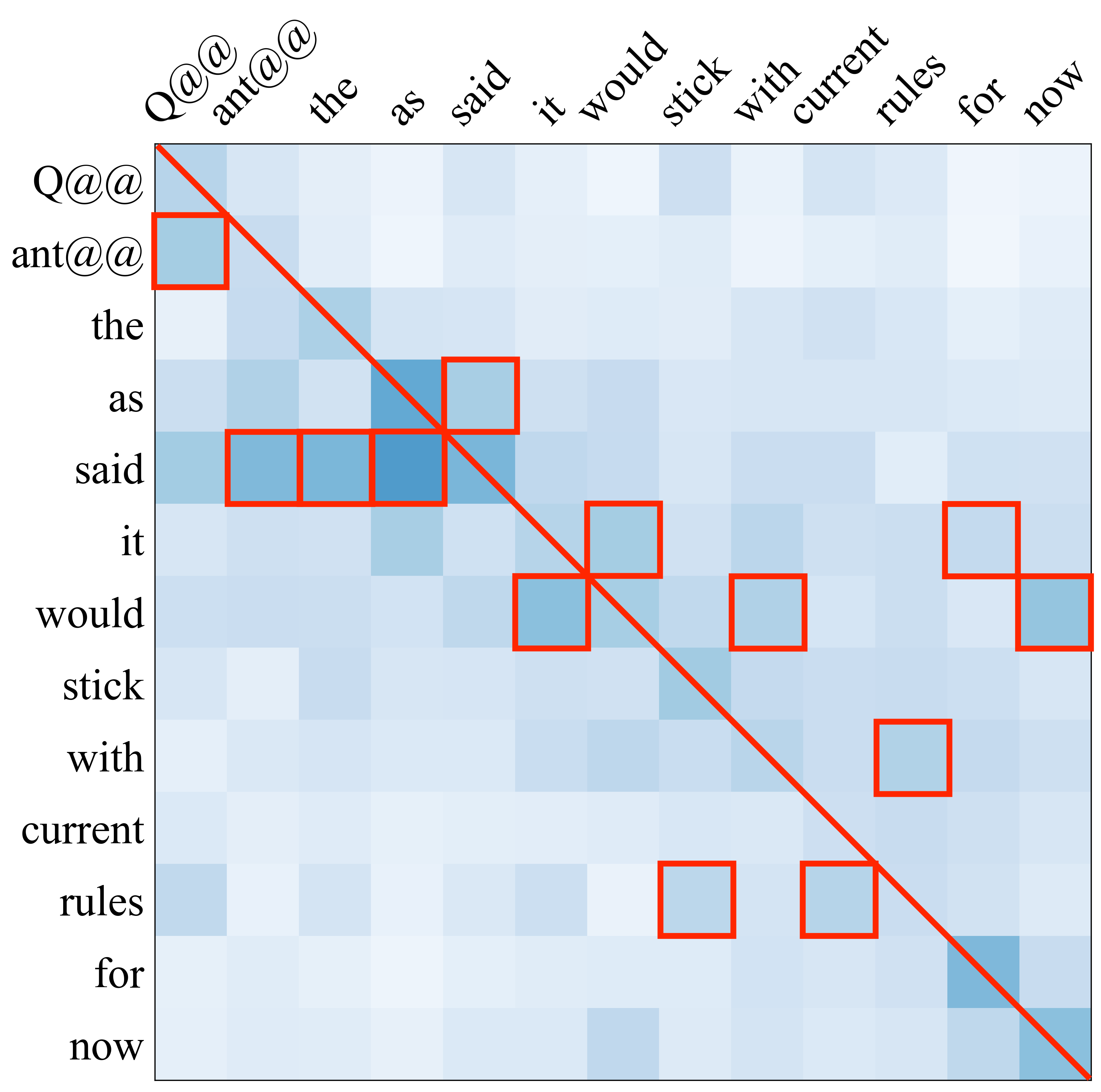}}
        \hfill
        \subfloat[SSANs]{\includegraphics[height=0.23\textwidth]{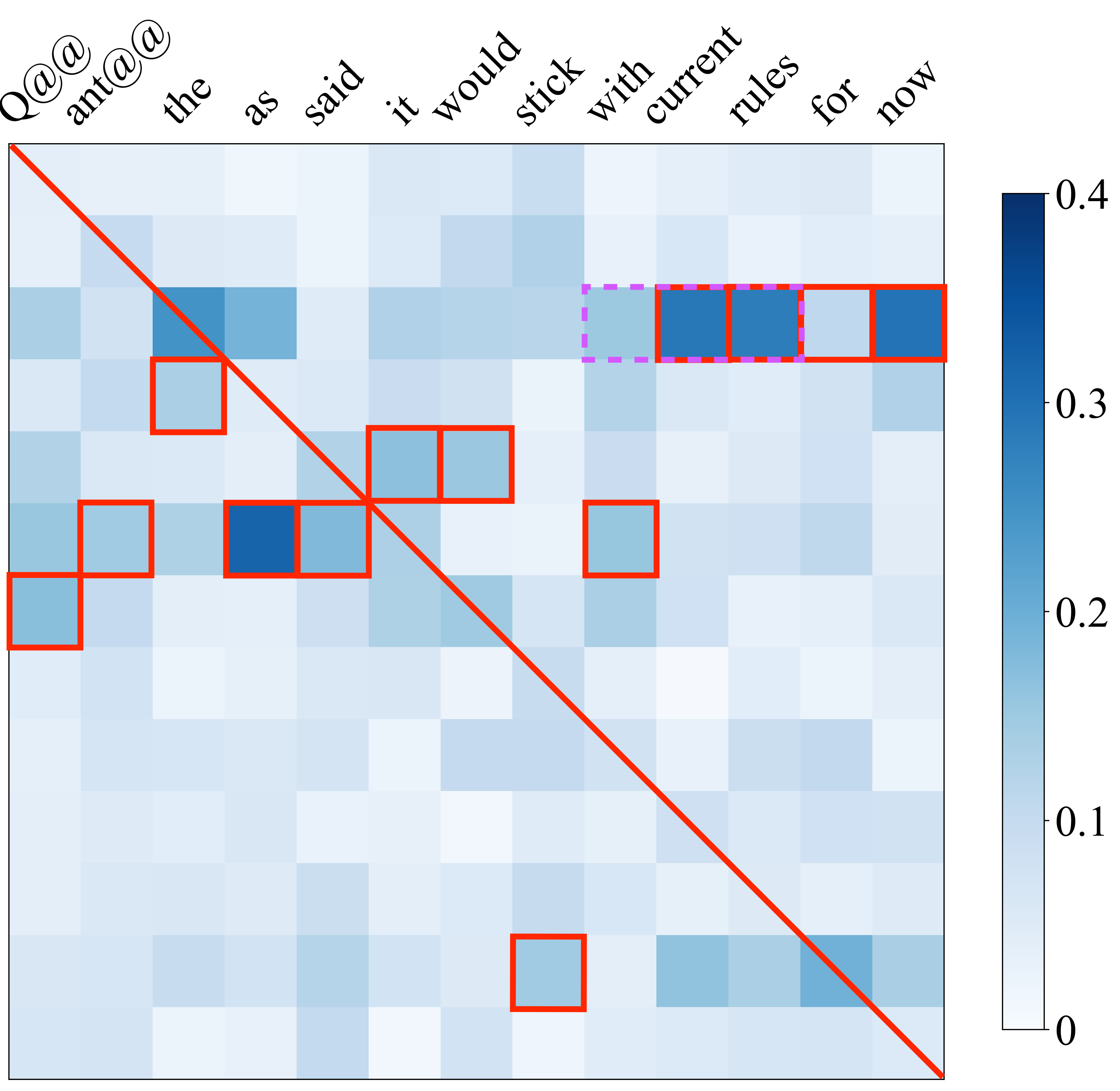}}
    \end{center}
	\caption{Visualization of attention weights from an example on the {\em global} reordering detection task. We highlight the attended word (Y-axis) with maximum attention weight for each query (X-axis) in red rectangles.}
	\label{fig:reorder:case}
\end{figure}

\paragraph{Attention Behaviors}
The objective of the WRD is to distinguish a global reordering (averaged distance is $8.7$ words), which requires the examined model to pay more attention to distant words.
Figure~\ref{fig:reorder:distance} depicts the attention distribution according to different relative distances.
SSANs alleviate the leaning-to-local nature of SANs and pay more attention to distant words (e.g., distance$>5$), which better accomplish the task of detecting global reordering.
Figure~\ref{fig:reorder:case} illustrates an example, in which more queries in SSANs attend most to the inserted word ``{\em the}'' than SANs. Particularly, SANs pay more attention to the surrounding words (e.g., distance $< 3$), while the inserted word ``{\em the}'' only accepts subtle attention. 
In contrast, SSANs dispense much attention over words centred on the inserted position (\ie ``{\em the}'') regardless of distance, especially for the queries ``{\em current rules for now}''.  We speculate that SSANs benefits from such features on detecting the global word reordering .

\section{Evaluation of Structure Modeling}
\label{sec:structure}

In this section, we investigate whether SSANs better capture structural information of sentences. To this end, we first empirically evaluate the syntactic structure knowledge embedded in the {\em learned representations} (\S\ref{sec:syntactic:probing}). Then we investigate the {\em attention behaviors} by extracting constituency tree from the attention distribution (\S\ref{sec:attention}).

\subsection{Structures Embedded in Representations}
\label{sec:syntactic:probing}

\begin{table}[t]
  \centering
  \begin{tabular}{c|r||cc|r}
     {\bf Class}  &  {\bf Ratio}  & {\bf SANs}  & {\bf SSANs} & \bm $\bigtriangleup$ \\ 
    \hline
    5    &  6.9\% & 68.66 & 75.22 & \em +9.6\%\\
    6    & 14.3\% & 56.10 & 64.09 & \em +14.2\%\\
    7    & 16.3\% & 46.63 & 55.05 & \em +18.1\%\\
    8    & 17.9\% & 39.68 & 50.88 & \em +28.2\%\\
    9    & 17.4\% & 38.33 & 50.97 & \em +33.0\% \\
    10   & 15.3\% & 35.54 & 49.88 & \em +40.3\% \\
    11   & 11.9\% & 48.86 & 56.39 & \em +15.4\%\\
    \hline
     All  & 100\% & 45.68 & 55.90 & \em +22.4\%\\ 
  \end{tabular}
  \caption{F1 score on the tree depth task. ``Ratio'' denotes the portion each class takes.}
  \label{tab:treedepth}
\end{table}

\begin{table}[t]
  \centering
  \begin{tabular}{c|r||cc|r}
     {\bf Type}  &  {\bf Ratio}  & {\bf SANs}  & {\bf SSANs} & \bm $\bigtriangleup$ \\ 
    \hline
    Ques. &  10\% & 95.90 & 97.06 & +1.2\%\\
    Decl. &  60\% & 88.48 & 91.34 & +3.2\%\\
    Clau. &  25\% & 72.78 & 78.32 &+7.6\%\\
    Other &  5\% & 50.67 & 61.13 & +20.6\% \\
    \hline
     All  & 100\% & 83.78 & 87.25 & \em +4.1\%\\ 
  \end{tabular}
  \caption{F1 score on the top constituent task. We report detailed results on 4 types of sentences: question (``Ques.''), declarative (``Decl.''), a clause (``Clau.''),nd other (``Other'') sentences.}
  \label{tab:topconst}
\end{table}

\paragraph{Task Description} 
We leverage two linguistic probing tasks to assessing the syntactic information embedded in a given representation. Both tasks are cast as multi-label classification problem based on the representation of a given sentence, which is produced by an examined model:

\noindent {\em Tree Depth} ({\bf TreeDepth}) task~\cite{conneau2018you} checks whether the examined model can group sentences by the depth of the longest path from root to any leaf in their parsing tree. Tree depth values range from 5 to 11, and the task is to categorize sentences into the class corresponding to their depth (7 classes).

\noindent {\em Top Constituent} ({\bf TopConst}) task~\cite{shi2016does} classifies the sentence in terms of the sequence of top constituents immediately below the root node, such as ``ADVP NP VP .''. The top constituent sequences fall into 20 categories: 19 classes for the most frequent top constructions, and one for all other constructions.

We trained the model on the open-source dataset provided by \newcite{conneau2018you}, and used the same model architecture in Section~\ref{sec:local-order}.

\paragraph{Probing Accuracy}
Table~\ref{tab:treedepth} lists the results on the TreeDepth task. SSANs significantly outperform SANs by 22.4\% on the overall performance. Concretely, the performance of SANs dramatically drops as the depth of the sentences increases.\footnote{The only exception is the class of ``11'', which we attribute to the extraction of feature of associating ``very complex sentence'' with maximum depth ``11''. } On the other hand, SSANs is more robust to the depth of the sentences, demonstrating the superiority of SSANs on capturing complex structures.

Table~\ref{tab:topconst} shows the results on the TopConst task. We categorize the 20 classes into 4 categories based on the types of sentences: question sentence (``* SQ .''), declarative sentence (``* NP VP *'' etc.), clause sentence (``SBAR *'' and ``S *''), and others (``OTHER''). Similarly, the performance of SANs drops as the complexity of sentence patterns increases (e.g., ``Ques.'' $\Rightarrow$ ``Others'', 95.90 $\Rightarrow$ 50.67). SSANs significantly improves the prediction F1 score as the complexity of sentences increases, which reconfirm the superiority of SSANs on capturing complex structures.

\subsection{Structures Modeled by Attention}
\label{sec:attention}

\begin{table}[t]
  \centering
  \begin{tabular}{c||cc|r}
	\bf Metric  &   \bf SANs    &   \bf SSANs   &   $\bigtriangleup$ \\
	\hline
	\bf BP   &   21.09   &   22.07   &  \em +4.7\%\\
	\bf BR      &   22.05   &   23.07   &  \em +4.6\%\\
	\hdashline
	\bf F1    &   21.56   &   22.56   &  \em +4.2\%\\
  \end{tabular}
  \caption{Evaluation on constituency trees generated from the attention distribution.}
  \label{tab:parsing}
\end{table}

\begin{figure*}[t]
\centering
\subfloat[SANs]{\includegraphics[width=0.46\textwidth]{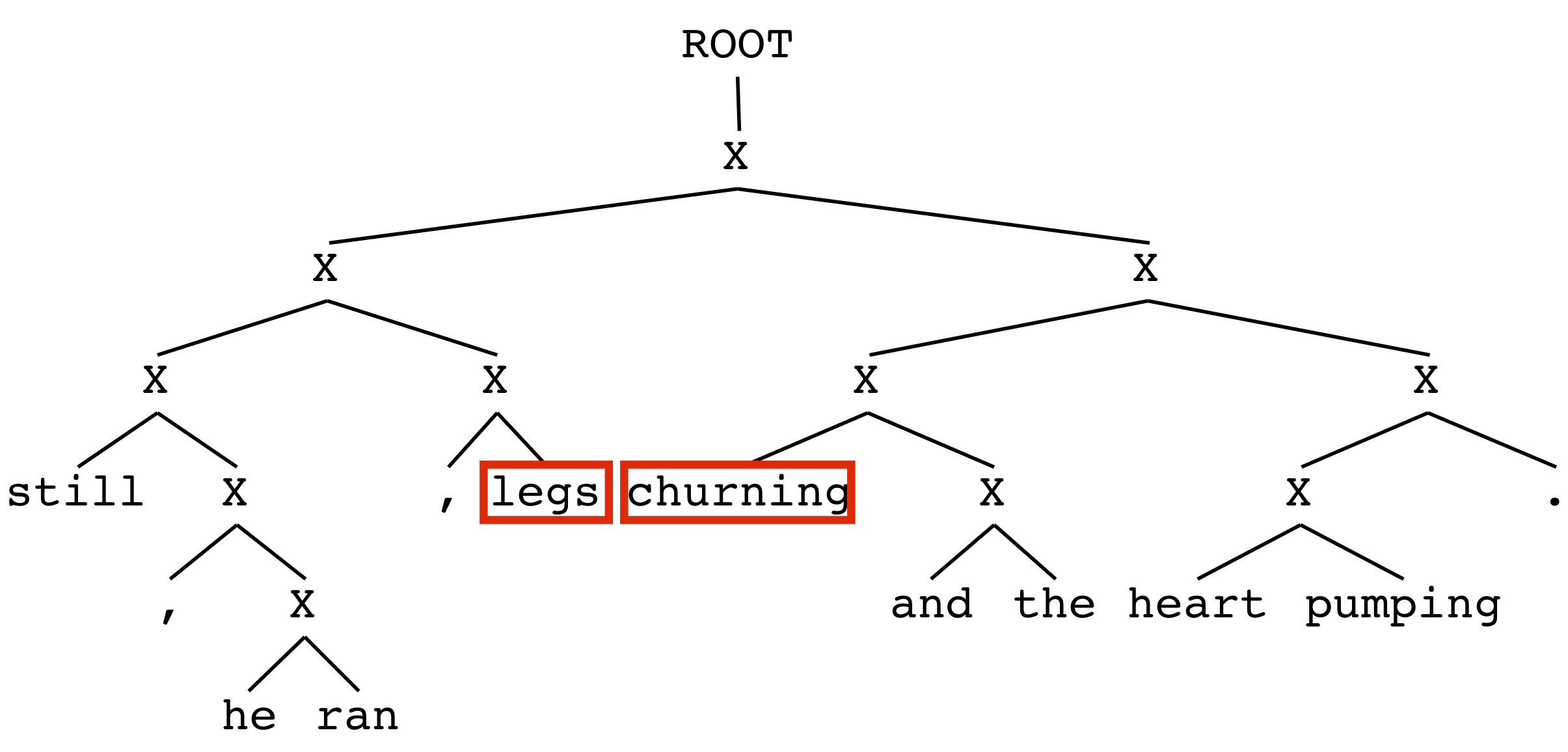}  \label{fig:tree-base}} \hfill
\subfloat[SSANs]{\includegraphics[width=0.46\textwidth]{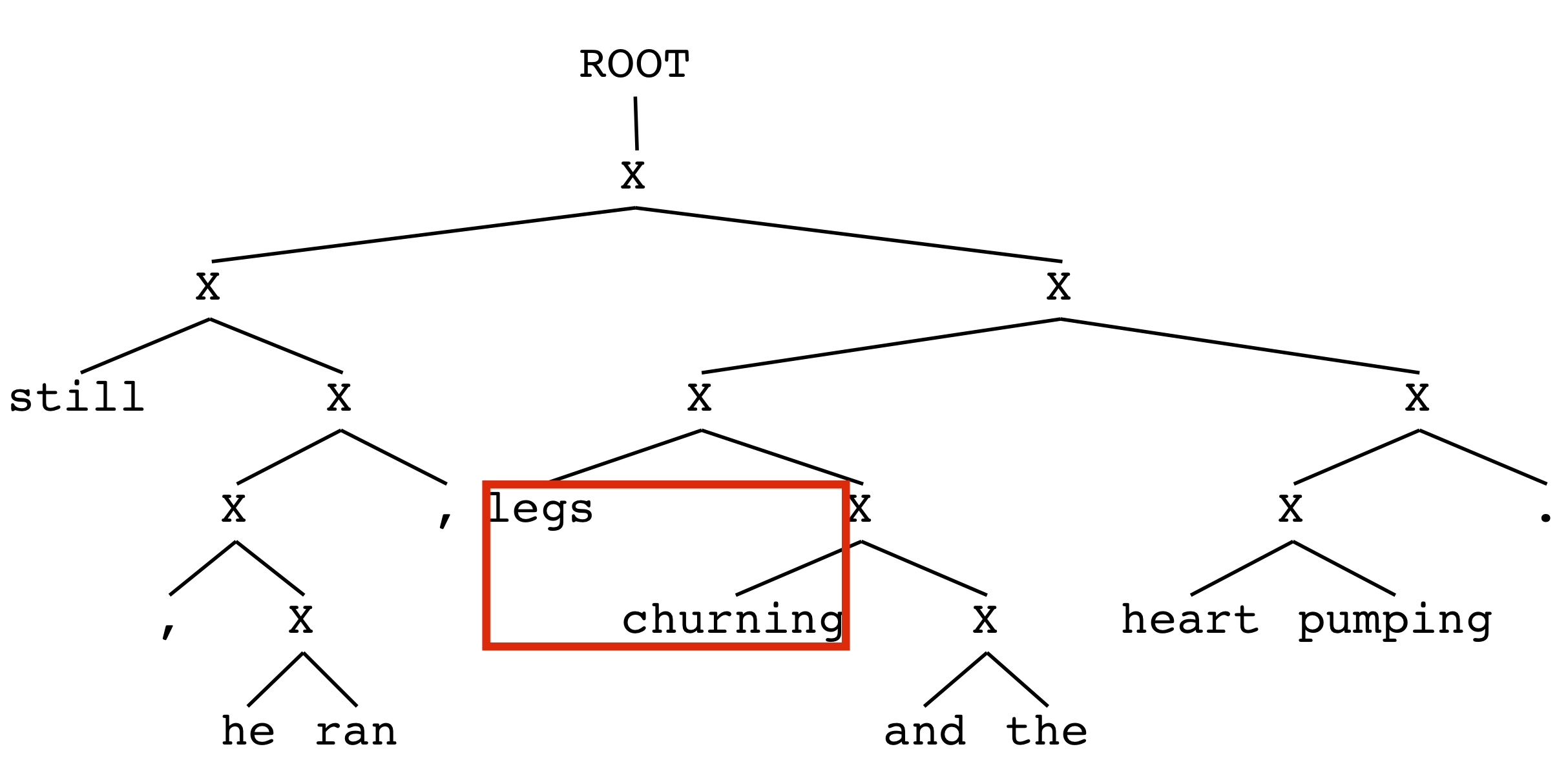} \label{fig:tree-ours}}
\caption{Example of constituency trees generated from the attention distributions.}
\label{fig:tree}
\end{figure*}

\begin{table*}[t]
  \centering
  \begin{tabular}{c|c||cc|r||cc|r||cc|r}
    \multicolumn{2}{c||}{\bf Type}  &  \multicolumn{3}{c||}{\bf TreeDepth}  &   \multicolumn{3}{c||}{\bf TopConst}  &   \multicolumn{3}{c}{\bf En$\Rightarrow$De Translation} \\
    \cline{3-11}
    \multicolumn{2}{c||}{}   & {\em SANs}  & {\em SSANs} & \em $\bigtriangleup$   & {\em SANs}  & {\em SSANs} & \em $\bigtriangleup$ & {\em SANs} & {\em SSANs} & \em $\bigtriangleup$\\ 
    \hline\hline
    \multirow{4}{*}{\rotatebox[origin=c]{90}{{\bf Content}}} & 
     Noun & 0.149 & 0.245 & \em +64.4\% & 0.126 & 0.196 & \em +55.6\%& 0.418 & 0.689 & \em +64.8\%\\ 
    & Verb & 0.165 & 0.190 & \em  +15.2\% & 0.165 &  0.201 & \em +21.8\%& 0.146 & 0.126 & \em -13.7\%\\
    & Adj. & 0.040 & 0.069 & \em +7.3\% & 0.033 & 0.054 & \em +63.6\%& 0.077 & 0.074 & \em -3.9\%\\
    \cdashline{2-11}
    & Total & 0.354 & 0.504 & \em +42.4\% & 0.324 & 0.451 & \em +39.2\% & 0.641 & 0.889 & \em +38.7\% \\
     \hline
    \multirow{5}{*}{\rotatebox[origin=c]{90}{{\bf Content-Free}}} 
    & Prep. & 0.135 & 0.082 & \em -39.3\% & 0.123 & 0.119 & \em -3.3\% & 0.089 & 0.032 & \em -64.0\%\\
    & Dete. & 0.180 & 0.122 & \em -32.2\% & 0.103 & 0.073 & \em -29.1\% & 0.070 & 0.010 & \em -85.7\%\\
    & Punc. & 0.073 & 0.068 & \em -6.8\% & 0.078 & 0.072 & \em -7.7\%& 0.098 & 0.013 & \em -86.7\%\\
    & Others & 0.258 & 0.224 & \em -13.2\% & 0.373 & 0.286& \em -23.3\% & 0.102 & 0.057& \em -41.1\%\\
    \cdashline{2-11}
    & Total & 0.646 & 0.496 & \em -23.3\% & 0.676 & 0.549 & \em -18.8\% & 0.359 & 0.111 & \em -69.1\% \\
  \end{tabular}
  \caption{Attention distributions on linguistic roles for the structure modeling probing tasks (\S\ref{sec:syntactic:probing}, ``TreeDepth'' and ``TopConst'') and the constituency tree generation task (\S\ref{sec:attention}, ``En$\Rightarrow$De Translation'').}
  \label{tab:linguistic}
\end{table*}

\paragraph{Task Description} We evaluate the ability of self-attention on structure modeling by constructing constituency trees from the attention distributions.  
Under the assumption that attention distribution within phrases is stronger than the other, \newcite{marecek-rosa-2018-extracting} define the score of a constituent with span from position $i$ to position $j$ as the attention merely inside the span denoted as $\text{score}(i,j)$. Based on these scores, a binary constituency tree is generated by recurrently splitting the sentence. When splitting a phrase with span $(i, j)$, the target is to look for a position $k$ maximizing the scores of the two resulting phrases:
\begin{equation}
    k = \argmax_{k^{'}}(\text{score}(i, k^{'})\cdot \text{score}(k^{'},j))
\end{equation}

We utilized Stanford CoreNLP toolkit to annotate English sentences as golden constituency trees. We used \verb|EVALB|\footnote{\url{http://nlp.cs.nyu.edu/evalb}.} to evaluate the generated constituency trees, including bracketing precision, bracketing recall, and bracketing F1 score.

\paragraph{Parsing Accuracy}
As shown in Table~\ref{tab:parsing}, SSANs consistently outperform SANs by 4.6\% in all the metrics, demonstrating that SSANs better model structures than SANs.
Figure~\ref{fig:tree} shows an example of generated trees. As seen, the phrases ``{\em he ran}'' and ``{\em heart pumping}'' can be well composed for both SANs and SSANS. 
However, SANs fail to parse the phrase structure ``{\em legs churning}'', which is correctly parsed by SSANs.

\subsection{Analysis on Linguistic Properties}
\label{sec:linguistic}

In this section, we follow~\newcite{He:2019:EMNLP} to analyze the linguistic characteristics of the attended words in the above structure modeling tasks, as listed in Table~\ref{tab:linguistic}. Larger relative increase (``$\bigtriangleup$'') denotes more attention assigned by SSANs.
Clearly, {\em SSANs pay more attention to content words in all cases}, although there are considerable differences among NLP tasks. 

Content words possess semantic content and contribute to the meaning of the sentence, which are essential in various NLP tasks. For example, the depth of constituency trees mainly relies on the nouns, while the modifiers (e.g., adjective and content-free words) generally make less contributions. The top constituents mainly consist of VP (95\% examples) and NP (75\% examples) categories, whose head words are verbs and nouns respectively. In machine translation, content words carry essential information, which should be fully transformed to the target side for producing adequate translations. Without explicit annotations, SANs are able to learn the required linguistic features, especially on the machine translation task (e.g., dominating attention on nouns). 
SSANs further enhance the strength by paying more attention to the content words. 

However, due to their high frequency with a limited vocabulary (e.g., 150 words\footnote{\url{https://en.wikipedia.org/wiki/Function\_word.}}), content-free words, or function words generally receive a lot of attention, although they have very little substantive meaning. This is more series in structure probing tasks (i.e., TreeDepth and TopConst), since the scalar guiding signal (i.e., class labels) for a whole sentence is non-informative as it does not necessarily preserve the picture about the intermediate syntactic structure of the sentence that is being generated for the prediction. On the other hand, the problem on content-free words is alleviated on machine translation tasks due to the informative sequence signals. 
SSANs can further alleviate this problem in all cases with a better modeling of structures. 

\section{Conclusion}

In this work, we make an early attempt to assess the strengths of the selective mechanism for SANs, which is implemented with a flexible Gumbel-Softmax approach. Through several well-designed experiments, we empirically reveal that the selective mechanism migrates two major weaknesses of SANs, namely {\em word order encoding} and {\em structure modeling}, which are essential for natural language understanding and generation.
Future directions include validating our findings on other SAN architectures (e.g., BERT~\cite{devlin2019bert}) and more general attention models~\cite{bahdanau2015neural,luong2015effective}.

\section*{Acknowledgments}
We thank the anonymous reviewers for their insightful comments. We also thank Xiaocheng Feng, Heng Gong, Zhangyin Feng, and Xiachong Feng for helpful discussion. This work was supported by the National Key R\&D Program of China via grant 2018YFB1005103 and National Natural Science Foundation of China (NSFC) via grant 61632011 and 61772156.

\nocite{feng:2016:english}
\nocite{feng:2018:topic}

\bibliography{acl2020}
\bibliographystyle{acl_natbib}

\end{document}